\documentclass[letterpaper, 10 pt, conference]{ieeeconf}  % Comment this line out if you need a4paper

\IEEEoverridecommandlockouts                              % This command is only needed if 
\usepackage{subfiles} 
\usepackage{bm}
\usepackage{algorithm}
\usepackage{caption}
\usepackage{subcaption}
\usepackage{algpseudocode}
\usepackage{graphicx}
\usepackage[T1, OT1]{fontenc}
\usepackage{xcolor}
\usepackage{mathtools}
\usepackage{array}
\usepackage{multirow}
\usepackage{colortbl}
\usepackage{pgfplots}
\pgfplotsset{compat=1.9}
\usepackage{pgfplotstable}
\usepackage{tikz}
\usepackage{soul}
\usepackage{hhline}

\makeatletter
\newcommand\resetstackedplots{
\makeatletter
\pgfplots@stacked@isfirstplottrue
\makeatother
\addplot [forget plot,draw=none] coordinates{(YOLOv5s,0) (YOLOv5l,0) };
}
\makeatother

\makeatletter
\let\NAT@parse\undefined
\makeatother

\usepackage{hyperref}

\newcommand{\specialcell}[2][c]{%
\begin{tabular}[#1]{@{}c@{}}#2\end{tabular}}

\title{\LARGE \bf
DualCam: A Novel Benchmark Dataset for Fine-grained Real-time Traffic Light Detection
}

\author{Harindu Jayarathne, Tharindu Samarakoon, Hasara Koralege, Asitha Divisekara, \\Ranga Rodrigo and Peshala Jayasekara% <-this % stops a space
\thanks{The authors are with the Department of Electronic and Telecommunication Engineering, University of Moratuwa, Moratuwa 10400, Sri Lanka. (E-mail: \{170258L, 170538V, 170407U, 170150A, ranga, peshala\}@uom.lk).}%
}
\begin{document}

\maketitle
\thispagestyle{empty}
\pagestyle{empty}

%%%%%%%%%%%%%%%%%%%%%%%%%%%%%%%%%%%%%%%%%%%%%%%%%%%%%%%%%%%%%%%%%%%%%%%%%%%%%%%%
\begin{abstract}

Traffic light detection is essential for self-driving cars to navigate safely in urban areas. Publicly available traffic light datasets are inadequate for the development of algorithms for detecting distant traffic lights that provide important navigation information. We introduce a novel benchmark traffic light dataset captured using a synchronized pair of narrow-angle and wide-angle cameras covering urban and semi-urban roads. We provide 1032 images for training and 813 synchronized image pairs for testing. Additionally, we provide synchronized video pairs for qualitative analysis. The dataset includes images of resolution 1920$\times$1080 covering 10 different classes. Furthermore, we propose a post-processing algorithm for combining outputs from the two cameras. Results show that our technique can strike a balance between speed and accuracy, compared to the conventional approach of using a single camera frame. The dataset is available at \url{https://github.com/harinduravin/DualCam}

% Traffic light detection importance

% research gap

% * Unavailability of synchronized narrow and wide images covering traffic lights in the distance.
% * Lack of chaotic conditions.

% xxxx images for training
% xxxx images for testing
% resolution of images

% results section and the control algorithm

\end{abstract}

%%%%%%%%%%%%%%%%%%%%%%%%%%%%%%%%%%%%%%%%%%%%%%%%%%%%%%%%%%%%%%%%%%%%%%%%%%%%%%%%

\section{INTRODUCTION}

% Autonomous driving has
For the rapid development of Advanced Driver Assistance Systems (ADAS) and Autonomous Driving Systems (ADS), traffic light and sign detection play a crucial role. Especially, traffic light detection and recognition are complex tasks due to the smaller object sizes, illumination variations, close resemblance with other objects and the dynamic nature at operating time. Detection in real-time can be even more challenging due to computational resource limitations.

Training and evaluation of detection algorithms on traffic light datasets is an important task.
Earlier, traffic light detection algorithms mainly depended on traditional image processing techniques \cite{lara, wangrobust, mapping,diazrobust}. Recent traffic light detectors \cite{deeptlr, bach2017, ssdtl, bach2018, grassman} are based on state-of-the-art object detectors such as Faster R-CNN \cite{ren2015faster}, YOLO \cite{ redmon2016you} and SSD \cite{liu2016ssd}. Several traffic light datasets are publicly available \cite{ lara,bstld,driveu,lisa}, but some of them are limited by image resolution, number of classes, number of annotations, and quality. Most of the existing datasets tackle traffic lights in the short-range distance only, limiting the capability of identifying traffic lights at high speed.

To tackle this issue, we present DualCam, a novel traffic light benchmark
dataset consisting of 2250 annotated images and 8321 object instances. The dataset covers 10 different traffic light classes. Our benchmark dataset consists of images generated from a pair of synchronized narrow-angle and wide-angle cameras. Such synchronization can assist traffic light detection from an extended range. This additional information will be valuable for the perception system of an ADAS/ADS to plan smooth navigation.

% Most of the datasets mentioned above tackle traffic lights in the short range distance. We provide a dataset of images generated from a narrow-angle camera and a wide-angle camera. We present DualCam, a novel traffic light benchmark dataset consisting of 2250 annotated images and 8321 object instances. The dataset covers 10 different traffic light classes. When synchronized, a pair of frames from the two cameras can provide traffic light information from an extended distance. This additional information will be valuable for the perception system of an ADAS to plan smooth navigation. 

% Most of the datasets mentioned above tackle traffic lights in the short range distance. We present DualCam, a novel traffic light benchmark dataset which consists of images generated from a pair of synchronized narrow-angle and wide-angle cameras, which can provide traffic light information from an extended range of distance. This additional information will be valuable for the perception system of an ADAS/ADS to plan smooth navigation. The dataset consists of 2250 annotated images and 8321 object instances under 10 different traffic light classes.

In order to combine detection results from the two camera frames, we propose a post processing algorithm that combines bounding boxes. The algorithm accurately combines all the detection results into the wide angle camera frame. Additionally, it suppresses double detection caused by the same object being detected simultaneously in both the frames. We test the algorithm on DualCam dataset by employing the YOLOv5 \cite{yolov5} object detector. Class-wise accuracy values and detection speeds are provided for comparison. Additionally, we summarize the effectiveness of combining results from two camera-frame outputs, compared to detection using individual frames.

\begin{figure}[t]
    \centering
    \begin{subfigure}[c]{0.49\columnwidth}
         \centering
         \includegraphics[width=\textwidth]{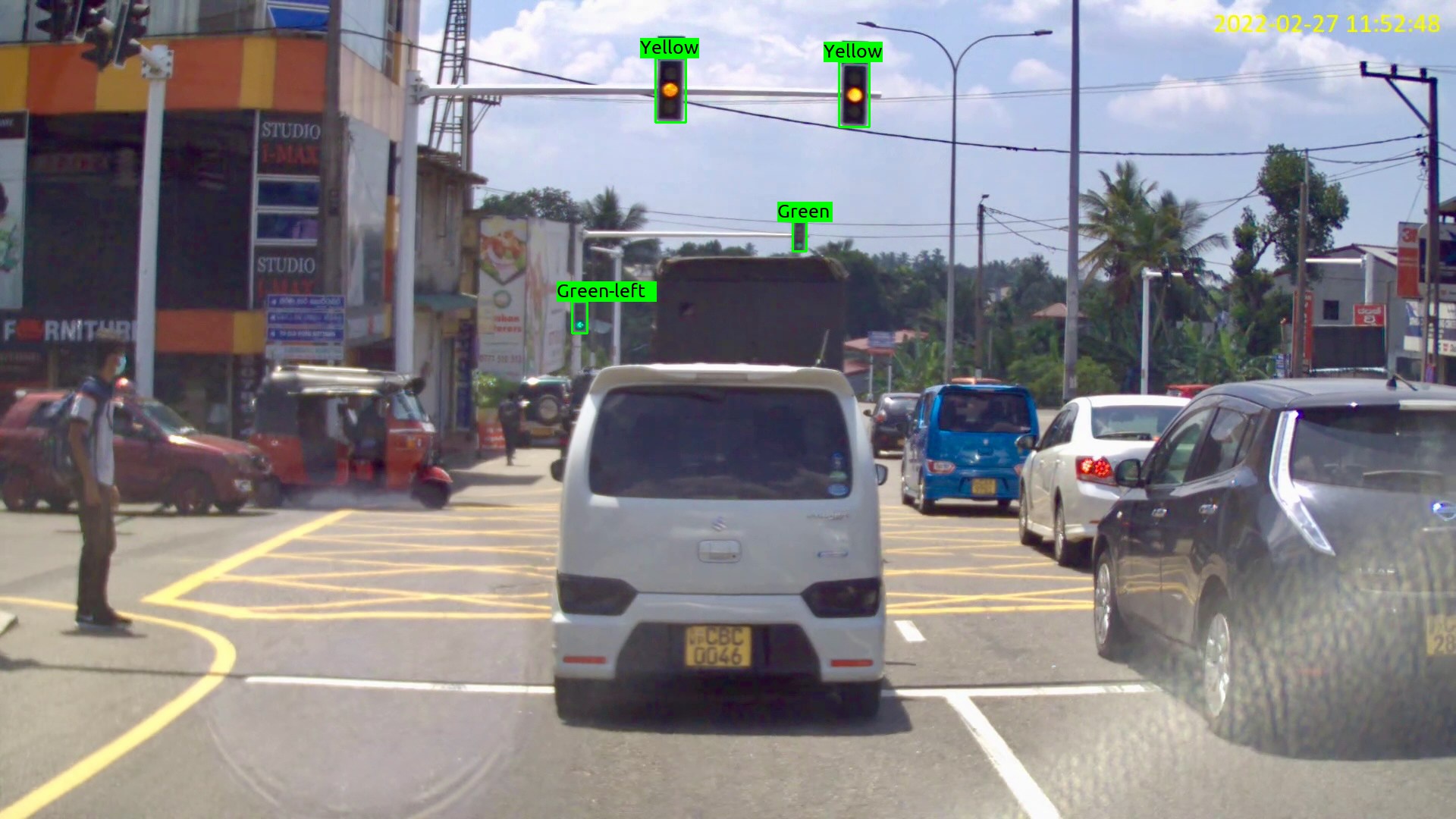}
         \caption{}
         \label{fig:narrowframe}
        % \vspace*{1mm}
     \end{subfigure}
     \hfill
    \begin{subfigure}[c]{0.49\columnwidth}
         \centering
         \includegraphics[width=\textwidth]{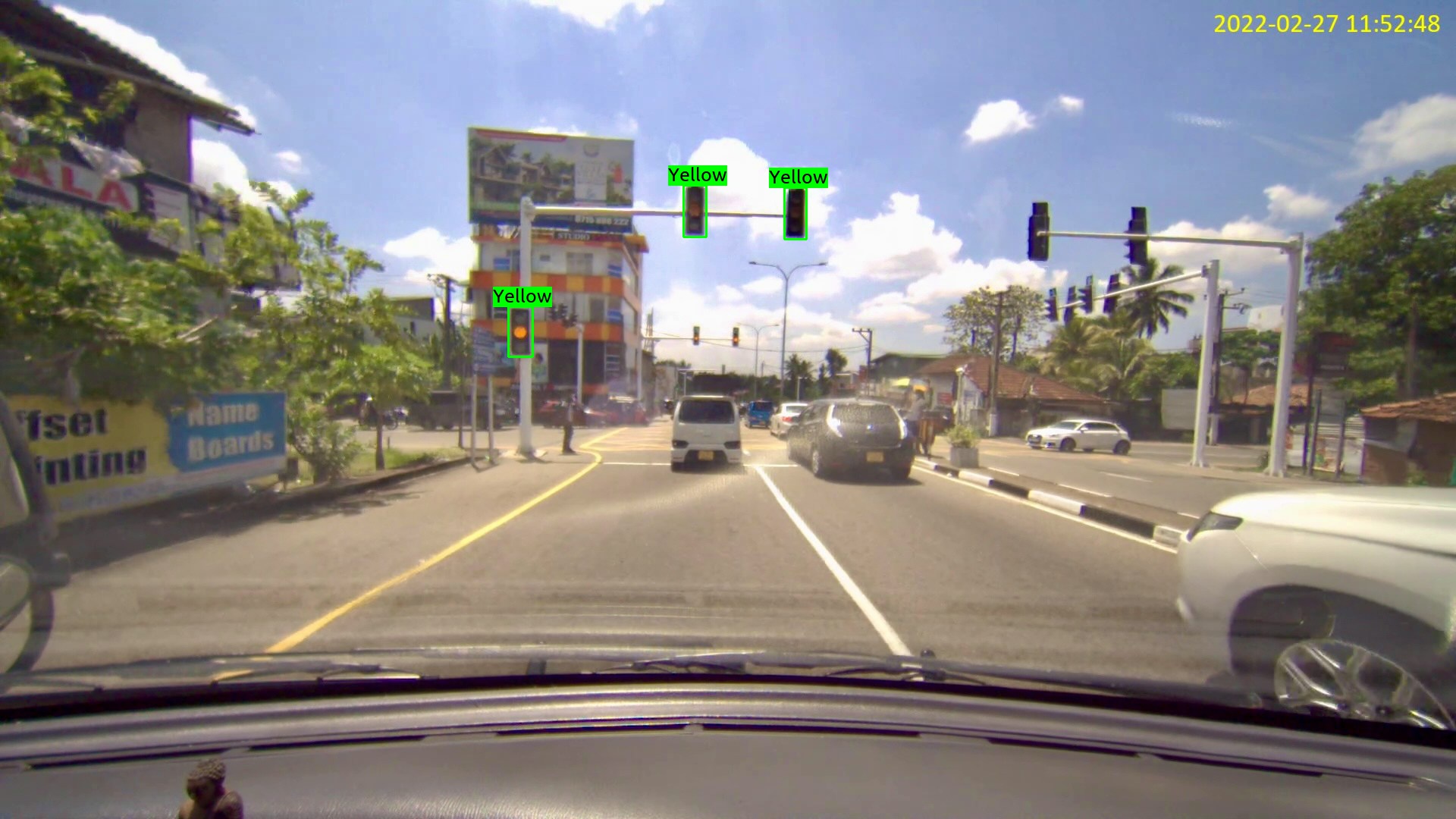}
         \caption{}
         \label{fig:wideframe}
        % \vspace*{0.1mm}
     \end{subfigure}
     \begin{subfigure}[c]{1\columnwidth}
         \centering
         \includegraphics[width=\textwidth]{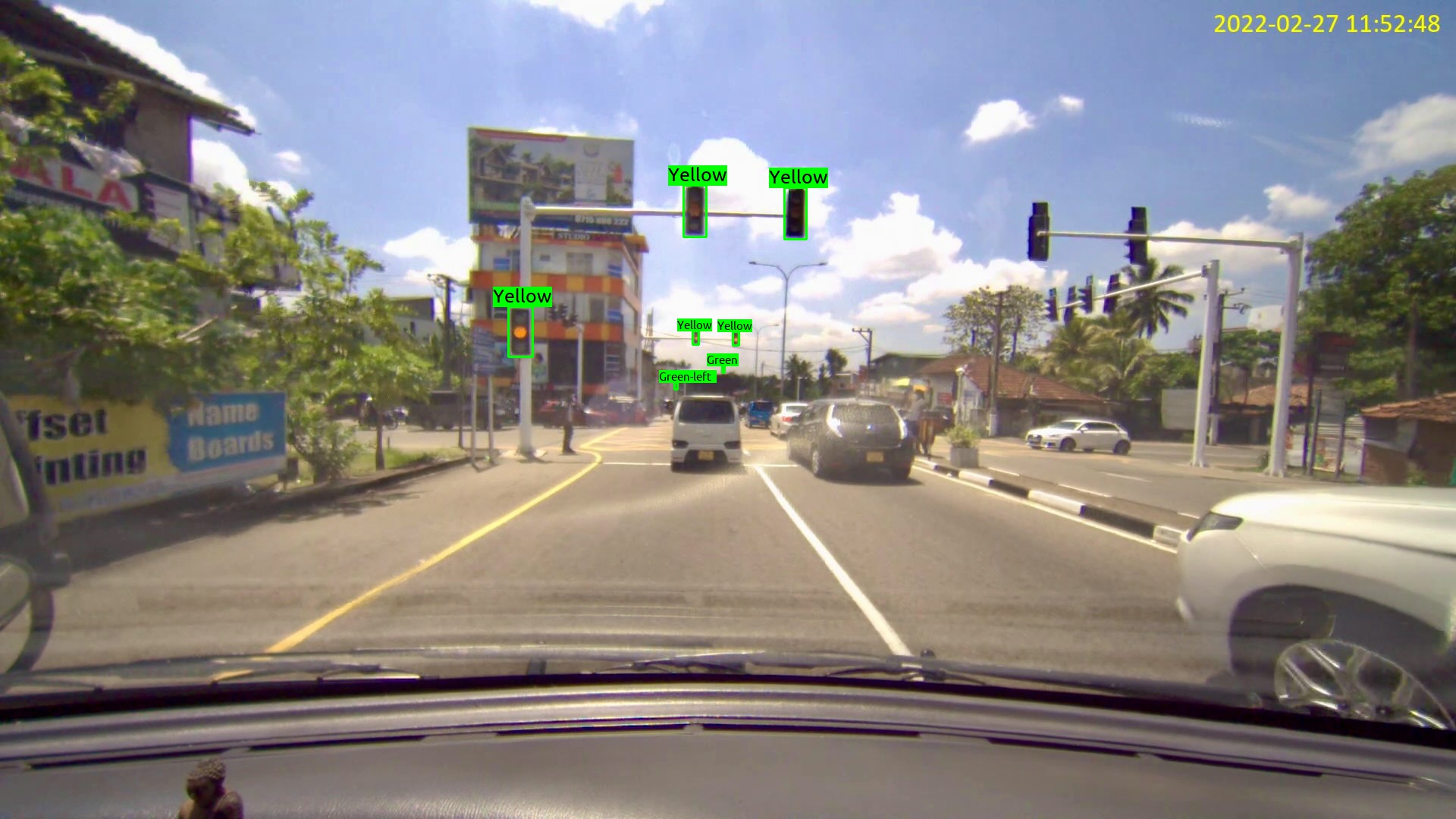}
         \caption{}
         \label{fig:integrated}
     \end{subfigure}
     \hfill
     
    \caption{A synchronized image pair from our dataset. (a) Narrow-angle camera frame. (b) Wide-angle camera frame. (c) Annotations from both frames integrated to wide-angle camera frame (Common ground truth).}
    \label{fig:lensesSensors}
    \vspace{-0.7cm}
\end{figure}

The use of multiple cameras for perception systems of self-driving cars is well-known. Stereo cameras are widely used to acquire depth information of driving scene. A traffic light dataset that is created using synchronized wide and narrow angle cameras will be a novel addition to the existing traffic light datasets.

Our contributions in this paper are as follows:
\begin{itemize}
  \item We provide a novel traffic light benchmark dataset covering 10 object classes, captured using a pair of cameras, one narrow-angled and the other wide-angled. A separate test set consisting of synchronized images and videos is provided for the purpose of testing algorithms with dual camera input.
  \item We propose a post-processing algorithm for real-time traffic light detection using a  synchronized narrow-angle and wide-angle camera pair. We discuss challenges that might arise in such a system.
    % \hl{detecting traffic lights using a narrow-angle and a wide-angle camera simultaneously in real-time.}
  \item We provide evaluation results of our benchmark dataset in terms of speed and accuracy using an object detection model. Results reveal that our approach strikes a balance between speed and accuracy.
\end{itemize}

This paper is organized as follows: First, we discuss related work in section II. We introduce our camera setup and the dataset in section III. A post processing algorithm for our dual camera setup is introduced in section IV. Section V includes the implementation details of our algorithm. The results of implementation are included in the Section VI.

\vspace{-0.1cm}

\section{RELATED WORK}\label{relatedwork}

\vspace{-0.1cm}

% In this section, we discuss different available datasets on traffic lights and related algorithms developed for the purpose of detecting traffic lights.
In this section, we discuss several publicly available traffic light datasets and traffic light detection algorithms.

\subsection{Datasets} 

After learning based detection methods became popular, the need for traffic light datasets increased. LARA dataset \cite{lara} provides frames of resolution $640 \times 480$ along with four different traffic light classes. The low resolution makes the dataset unsuitable for accurate detection. LISA dataset \cite{lisa}, which has been captured using stereo vision cameras, provides 14386 annotated daytime images with 7 traffic light classes. Even though the number of frames is high, the diversity of frames available is low. The frames have been captured at close time intervals, resulting in nearly identical frames.

% \textcolor{red}{para content - small traffic light detection datasets??}

More recently introduced Bosch Small Traffic Light Dataset (BSTLD) \cite{bstld} provides 13334 frames of resolution $1280 \times 720$ with detailed annotations up to one pixel width bounding boxes. The dataset provides 13 traffic light classes for training and 4 traffic light classes for testing. The class frequency is skewed towards most common traffic light states of red, yellow, green and empty. The DriveU Traffic Light Dataset \cite{driveu} is a large dataset that provides 230,000 annotations with wide range of traffic light classes. Due to their proposed class attribute structure, traffic lights can be divided into 344 unique classes. It provides an overview of existing datasets along with evaluation metrics for comparison purposes.

State-of-the-art datasets such as COCO \cite{lin2014microsoft} and Cityscapes \cite{cordts2016cityscapes} datasets include traffic lights as an annotation class. Even though traffic light detectors can be trained using these datasets, they cannot be trained for the classification task.

\subsection{Traffic light detection algorithms} 

Initially traffic light detection systems were mainly based on image processing and machine learning techniques. Under image processing techniques, transformation to different color spaces and subsequent thresholding \cite{lara,wangrobust} were used for the identification of potential regions of traffic lights. Further filtering using shape filters \cite{diazrobust} and template matching \cite{lara,wangrobust} were also common among the detection algorithms. Machine learning based algorithms such as support vector machine (SVM) \cite{mapping}, tree-based models \cite{lisa} use features such as histogram of gradients (HOG) for classification. Some methods such as \cite{mapping} need prior knowledge about the locations of traffic lights. Although these methods are computationally less expensive, detection is not accurate in complex scenarios.

%%%%%%%%%%%%%%%%%%%%%%%%%%%%%%%%%%%%%%%%

Recent deep learning object detection approaches out-perform all classical methods. Most of the methods adopt and customize state-of-the-art object detection algorithms. The most initial work that uses deep learning approaches for traffic light detection is DeepTLR \cite{deeptlr}. It introduces a convolutional neural network (CNN) that creates a pixel-wise probability map followed by a bounding box regressor. Traffic light detection algorithms introduced in \cite{bach2017,muller2017} consist of multi-camera systems for accurate detection from far distance. They have demonstrated the performance improvement by utilizing multiple-camera combinations. However, the algorithms presented are not compatible with modern object detectors.

A YOLO \cite{redmon2016you} based traffic light detection algorithm is introduced in \cite{bstld}. A separate custom classifier is used for the purpose of identifying traffic light status. Additionally, a stereo vision based object tracker is introduced. Traffic light detection algorithms in \cite{bach2018,grassman} are based on customized versions of Faster-RCNN \cite{ren2015faster} and an adapted version of SSD \cite{liu2016ssd} is used in \cite{ssdtl}. Even though these algorithms are robust towards detecting smaller traffic lights, the speed of the algorithms are not suitable for resource-constrained real-time applications.

% Traffic light detection algorithms in \cite{bach2018,grassman} and \cite{ssdtl} are based on customized versions of Faster-RCNN \cite{ren2015faster} and SSD \cite{liu2016ssd}, respectively.

% Traffic light detection algorithms in \cite{bach2018,grassman} use Faster-RCNN \cite{ren2015faster}, whereas \cite{ssdtl} uses SSD \cite{liu2016ssd}.

% \hl{Authors of} \cite{grassman} uses Grassman Manifold learning based technique for better recognition of traffic light state.

% 1) Deep TLR - 2016
% 2) Bach - 2017 - camera fusion
% 3) BSTLD - 2017
% 4) Bach - 2-18 - Faster RCNN
% 4) TL - SSD - 2018
% 5) Grassman manifold - 2019
% 6)  Changsuk - 2020

\section{DUALCAM TRAFFIC LIGHT DATASET}\label{dataset}

This section covers the details about the DualCam traffic light dataset including data collection details and dataset statistics.

\subsection{Data collection and annotation} 

\subsubsection{Camera details}

\begin{figure}[!h]
    \centering
    \begin{subfigure}[c]{0.65\columnwidth}
         \centering
         \includegraphics[width=\textwidth]{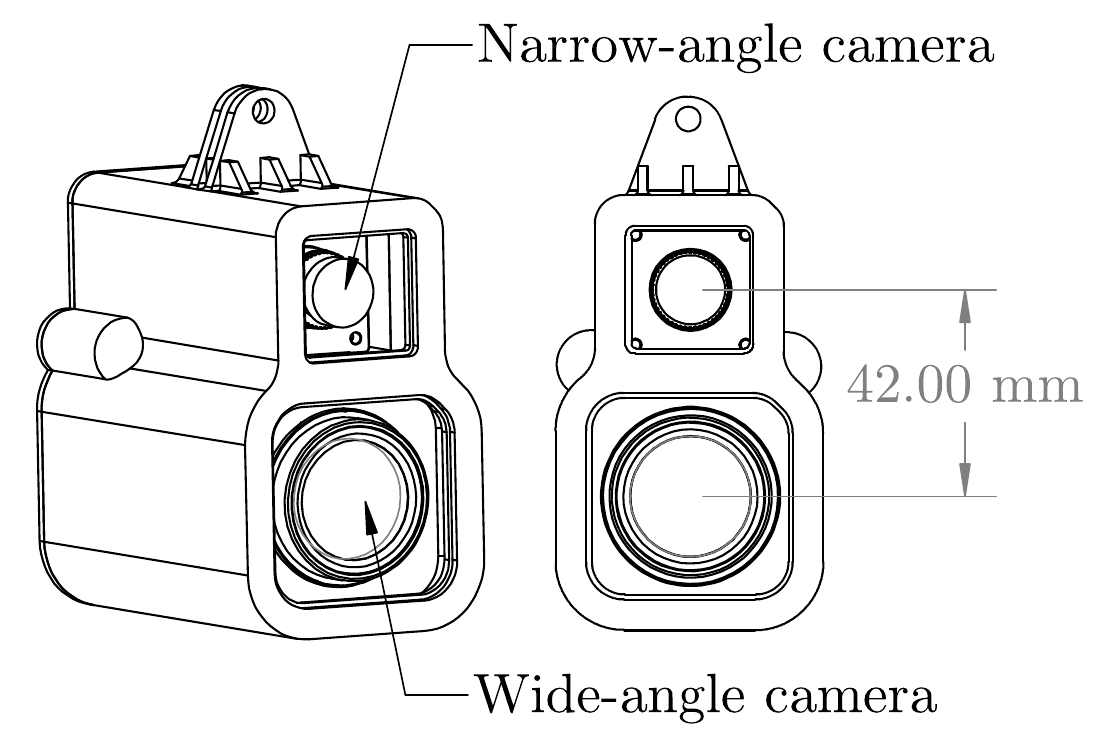}
         \caption{}
        %  \label{fig:y equals x}
        % \vspace*{1mm}
     \end{subfigure}
     \hfill
    \begin{subfigure}[c]{0.3\columnwidth}
         \centering
         \includegraphics[width=\textwidth]{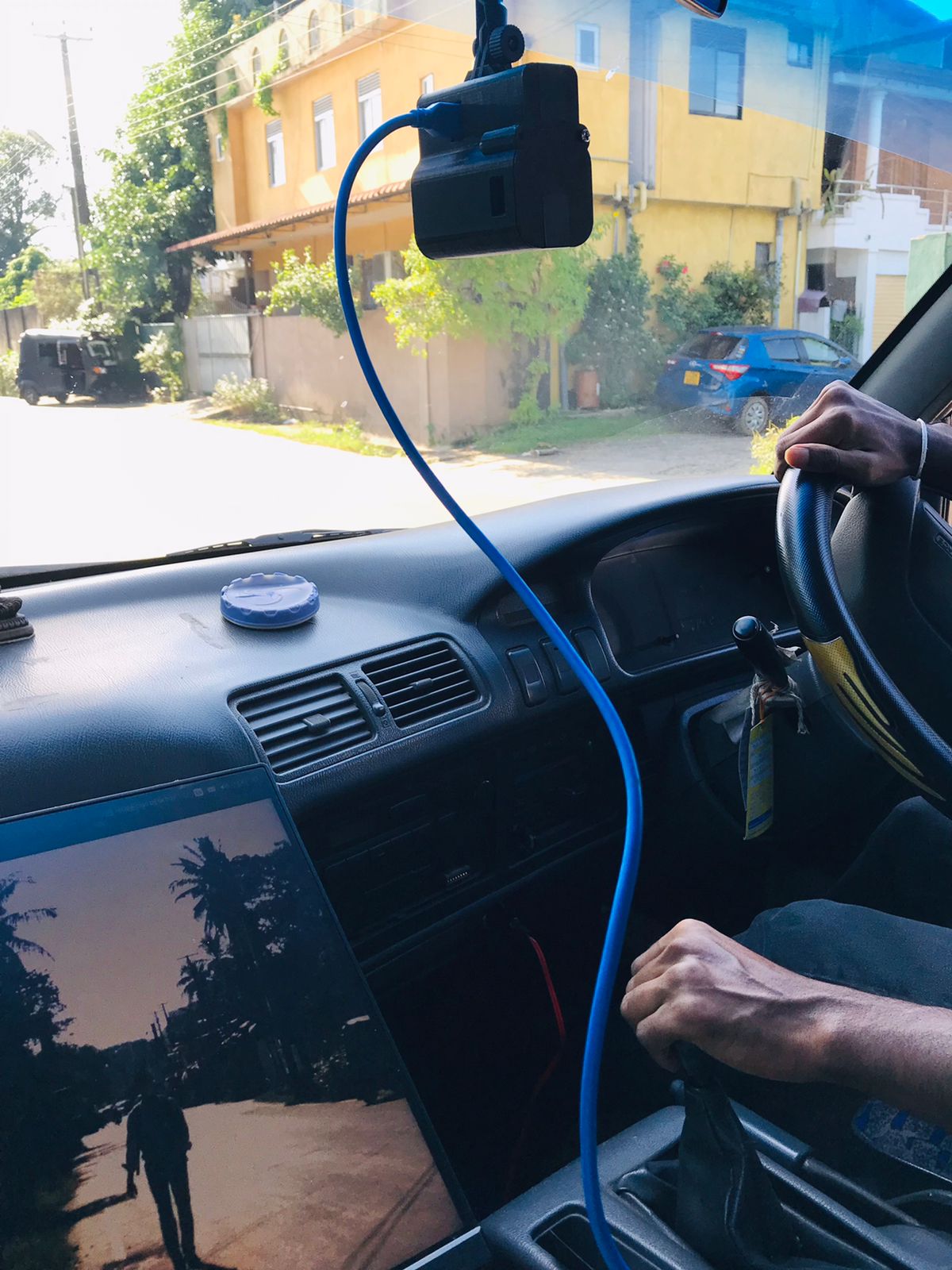}
         \caption{}
        %  \label{fig:y equals x}
        % \vspace*{0.1mm}
     \end{subfigure}

    \caption{Camera system. (a) Vertically aligned wide-angle and narrow-angle cameras in a single enclosure. (b) Camera system is mounted behind the windshield of the vehicle.}
    \label{fig:camera_system}
\end{figure}

Our camera system consists of two vision-grade Basler daA1920-30uc USB cameras with 1920$\times$1080 resolution: a narrow-angle camera with $48^0$ horizontal field of view (FoV) and a wide-angle camera with $125^0$ horizontal FoV. They are vertically aligned in a single enclosure mounted behind the windshield of the vehicle (Figure \ref{fig:camera_system}). The baseline between two cameras is 42mm. The narrow-angle camera is mainly used to detect the traffic lights located far away from the vehicle. The wide-angle camera is used to detect the traffic lights nearby the vehicle. The cameras provide synchronized image pairs. 
% More details about the camera system are given in the table \ref{tab:cam_specs}. 
% (The pixel size is 2.2um).  Further, details about cameras are given in the table (no. …) 

% Vertically aligned wide-angle and narrow-angle cameras in a tight enclosure.

% \begin{figure}[!h]
%     \centering
%     \includegraphics[width=0.4\textwidth]{images/camera_dimensions.pdf}
%     \caption{Camera system}
%     \label{fig:camera_system}
% \end{figure}

% \begin{table}[!h]
%     \caption{Camera specifications - \textcolor{red}{ida nathi nisa ain krmu}}
%     \centering
%     \scriptsize
%     \begin{tabular}{|m{2.5cm}|m{1.4cm}|m{0.8cm}|m{1.8cm}| } \hline
%          & \textbf{Focal Length} & \textbf{VFoV} & \textbf{Frame Rate (fps)} \\ \hline
%         \textbf{Wide angle camera} & 1.3mm & $109^0$ & 30 \\ \hline
%         \textbf{Narrow angle camera} & 6mm & $34.5^0$ & 30 \\ \hline
%     \end{tabular}

%     \label{tab:cam_specs}
% \end{table}

\subsubsection{Procedure}

The data collection was carried out in semi-urban and urban areas, where the density of traffic lights is high. Footage from both cameras are recorded at a frequency of 30 FPS while driving at regular speed. Image frames are extracted from the footage and manually annotated using CVAT \cite{cvat} annotation tool. We present our dataset in both PASCAL VOC XML \cite{pascalvoc} and YOLO \cite{ redmon2016you} annotation format.

\subsection{Dataset statistics}

DualCam dataset is divided into training and test sets.
The training set contains 1032 images in total; out of them, 776 images are from the narrow-angle camera and the rest of the 256 images are from the wide-angle camera. The test set contains 1626 images. It consists of 813 image pairs captured simultaneously using the two cameras. The test set is larger compared to conventional datasets, since we provide additional synchronized image pairs for better evaluation. Additionally, 40 minutes of synchronized test video pairs from the two cameras are provided. The dataset contains 10 object classes related to traffic light detection. The frequency of each class is provided in the Table \ref{table:instance_count}.
% two sets of images from narrow-angle camera and wide-angle camera. ( How many? 1200- narrow, 500- wide)

% The detection model can be trained using both types of images.

% The test set consist of images and videos. The images are synchronized image pairs captured from narrow-angle and wide-angle cameras (How many image pairs- 300 ). 

% Videos are also provided as pairs that can be used for testing the detection algorithm. (How many pairs - 10 - how long)

\begin{table}[ht]
\caption{Number of instances in the classes. The letters in parentheses denote the corresponding labels out of images in the Figure \ref{fig:traffic_lights}}

\begin{center}
\begin{tabular}{|l|c|c|c|}
\hline
\cellcolor{gray!15}{\textbf{Traffic light class}} & \cellcolor{gray!15}{\textbf{Train set}} & \cellcolor{gray!15}{\textbf{Test set}} & \cellcolor{gray!15}{\textbf{Total}}\\
\hline
Green \textcolor{gray}{(d)}           	& 1198	    & 1251	    & 2449  \\ \hline
Red \textcolor{gray}{(a)}             	& 565	    & 901	    & 1466  \\ \hline
Green-up \textcolor{gray}{(f)}        	& 426	    & 495	    & 921   \\ \hline
Empty-count-down \textcolor{gray}{(j)}	& 537	    & 225	    & 762   \\ \hline
Count-down \textcolor{gray}{(i)}      	& 346	    & 396	    & 742   \\ \hline
Yellow \textcolor{gray}{(c)}          	& 452	    & 246	    & 698   \\ \hline
Empty \textcolor{gray}{(h)}           	& 222	    & 469	    & 691   \\ \hline
Green-right \textcolor{gray}{(g)}     	& 115	    & 171	    & 286   \\ \hline
Green-left \textcolor{gray}{(e)}      	& 55	    & 105	    & 160   \\ \hline
Red-yellow \textcolor{gray}{(b)}      	& 66	    & 80	    & 146   \\ \hline
\multicolumn{1}{c|}{} & \textbf{3982} & \textbf{4339} & \textbf{8321} \\ \cline{2-4}
\end{tabular}
\end{center}
\label{table:instance_count}
\end{table}

\begin{figure}[!h]
    \vspace{0.2cm}
    \centering
    \begin{subfigure}[b]{0.088\columnwidth}
         \centering
         \includegraphics[width=\textwidth]{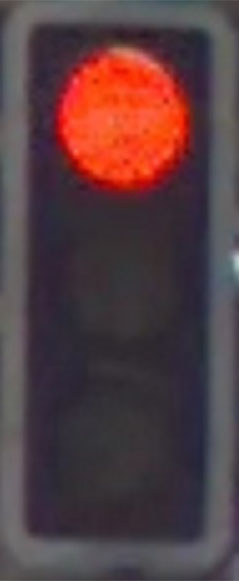}
         \caption{}
     \end{subfigure}
     \hfill
    \begin{subfigure}[b]{0.088\columnwidth}
         \centering
         \includegraphics[width=\textwidth]{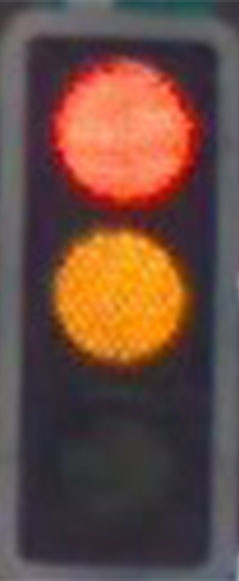}
         \caption{}
     \end{subfigure}
     \hfill
    \begin{subfigure}[b]{0.088\columnwidth}
         \centering
         \includegraphics[width=\textwidth]{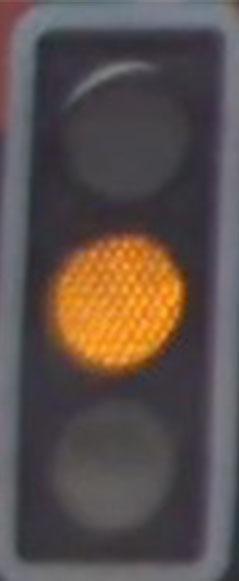}
         \caption{}
     \end{subfigure}
    \begin{subfigure}[b]{0.088\columnwidth}
         \centering
         \includegraphics[width=\textwidth]{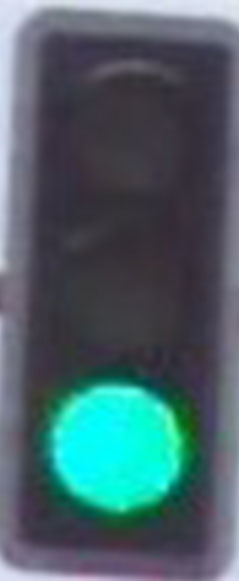}
         \caption{}
     \end{subfigure}
    \begin{subfigure}[b]{0.088\columnwidth}
         \centering
         \includegraphics[width=\textwidth]{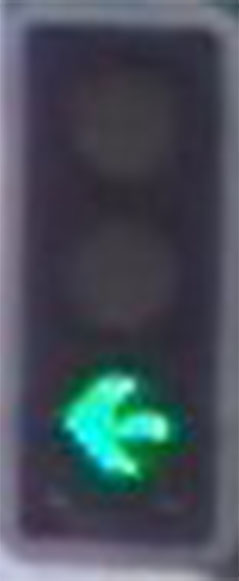}
         \caption{}
     \end{subfigure}
     \hfill
    \begin{subfigure}[b]{0.088\columnwidth}
         \centering
         \includegraphics[width=\textwidth]{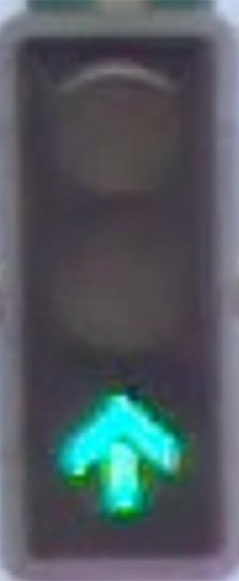}
         \caption{}
     \end{subfigure}
    \begin{subfigure}[b]{0.088\columnwidth}
         \centering
         \includegraphics[width=\textwidth]{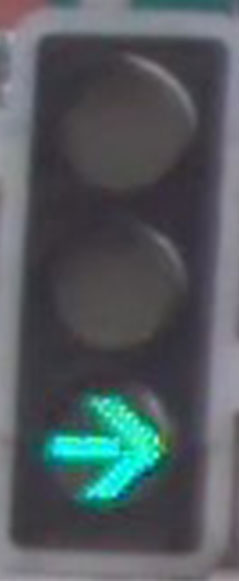}
         \caption{}
     \end{subfigure}
     \hfill
    \begin{subfigure}[b]{0.088\columnwidth}
         \centering
         \includegraphics[width=\textwidth]{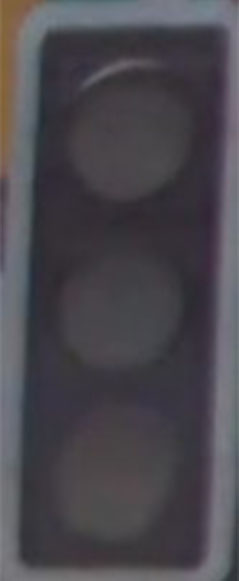}
         \caption{}
     \end{subfigure}
    \begin{subfigure}[b]{0.088\columnwidth}
         \centering
         \includegraphics[width=\textwidth]{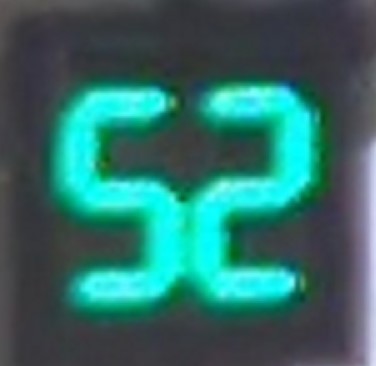}
         \caption{}
     \end{subfigure}
     \hfill
    \begin{subfigure}[b]{0.088\columnwidth}
         \centering
         \includegraphics[width=\textwidth]{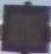}
         \caption{}
     \end{subfigure}

    \caption{Ten traffic light classes available in the dataset. (a) red, (b) red-yellow, (c) yellow, (d) green, (e) green-left, (f) green-up, (g) green-right, (h) empty, (i) count-down, (j) empty-count-down.}
    \label{fig:traffic_lights}

\end{figure}

\section{POST PROCESSING ALGORITHM}\label{algorithm}

% Our detection algorithm is based on YOLOv5 object detector. In order to tailor the algorithm towards detecting traffic lights, we propose following approach. 

At a given moment the two cameras can produce two image frames simultaneously. These two frames are fed to the object detector as a single batch of size two. The bounding boxes of detected traffic lights in two frames should be concatenated to obtain useful results. As wide-angle camera frame has wider field of view, we transform detected bounding boxes in narrow-angle camera frame (Figure \ref{fig:narrowframe}) to match with the wide-angle camera frame. Then we concatenate all the detected bounding boxes together within wide-angle camera frame (Figure \ref{fig:integrated}). The bounding box transformation and concatenation are the major steps in this approach.

\subsection{Bounding box transformation}\label{box_tran}
 
% We do the inferencing on original (distorted) camera frames and do necessary transformations on the detected bounding boxes to concatenate them as mentioned previously. This method avoids transforming the full images and reduces required computational power. It is favorable towards real-time traffic light detection.

% The two cameras are in a single fixed enclosure. Therefore, the relationship between two cameras is fixed. We used this relationship to transfer detected bounding boxes in original (undistorted) narrow angle camera frame to original (undistorted) wide-angle camera frame. 
 
\begin{figure*}
    \centering
    \includegraphics[width = 1\textwidth]{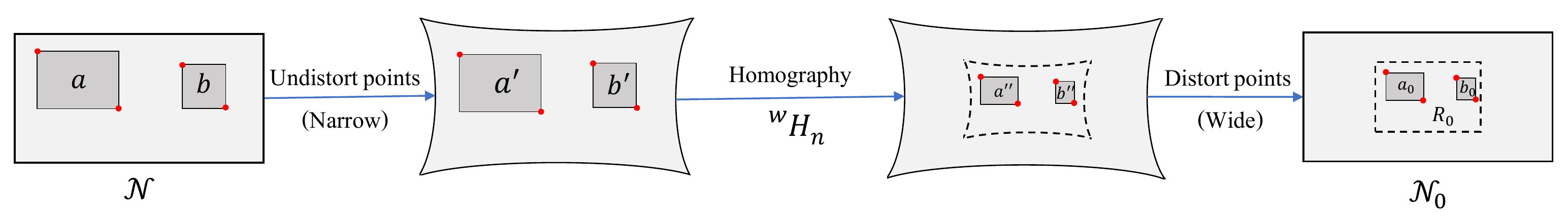}
    \caption{Bounding box transformation steps. First, boxes in original (distorted) narrow-angle camera frame ($a,b$) are mapped to undistorted narrow-angle camera frame ($a^{'},b^{'}$) using camera distortion parameters. Then boxes are mapped to undistorted wide-angle camera frame ($a^{''},b^{''}$) using planar homography. Finally, boxes are mapped to original (distorted) wide-angle camera frame ($a_0,b_0$) using its distortion parameters.}
    \label{fig:bboxtran}
\end{figure*}

Bounding box transformation is done in three steps as shown in Figure \ref{fig:bboxtran}. First, the bounding boxes detected in original (distorted) narrow-angle camera frame are transformed into its undistorted camera frame using camera distortion parameters. Then these boxes are transformed into undistorted wide-angle camera frame using planar homography between the two camera frames. Finally, they are transformed on to original (distorted) wide-angle camera frame using camera distortion parameters. Details on homography matrix estimation can be found in section \ref{homography_estimation}. Compared to transforming full images, transforming the bounding box coordinates requires less computation power.

\subsection{Bounding box concatenation}\label{box_concat}
 
%%%%%%%%%%%%% 

The bounding boxes of detected traffic lights in narrow-angle camera frame are transformed and concatenated with the bounding boxes of detected traffic lights in the wide-angle camera frame. Some traffic lights detected in the wide-angle camera frame might be detected fully or partially in the narrow-angle camera frame creating duplicate bounding boxes. We remove these duplicate bounding boxes as described below.
 
Let the sets of bounding boxes generated by narrow-angle camera and wide-angle camera be denoted as $\mathcal{N}$ and $\mathcal{W}$. Let the bounding boxes of $\mathcal{N}$ transformed using the transformation technique mentioned in the section \ref{box_tran} be denoted as $\mathcal{N}_{0}$. All $w \in \mathcal{W}$ that reside completely within the region where the narrow angle camera frame is mapped (called $R_0$) are removed from the set $\mathcal{W}$. The resulting set is $\mathcal{W}_r$.
% First if a bounding box in wide angle camera frame is fully inside the region where the narrow angle camera frame is mapped (called $R_0$), it is removed.
 
To handle the bounding boxes generated from the wide-angle camera that reside partially inside $R_0$, let $q \in \mathcal{Q}$ denote the shapes generated by taking the intersection between $w_r$ and $R_0$ for all $w_r \in \mathcal{W}_r$. An associated bounding box in $\mathcal{N}_{0}$ can exist for all $q \in \mathcal{Q}$. The association is measured using the Intersection over Union (IoU). If the IoU calculated between each $q \in \mathcal{Q}$ and $n_0 \in \mathcal{N}_0$ exceeds a certain threshold $\zeta$, the corresponding bounding box is removed from $\mathcal{N}_0$ to create the new set $\mathcal{N}_r$. Finally, bounding boxes from $\mathcal{N}_r \cup \mathcal{W}_r$ are taken onto the wide-angle camera frame as the output. The steps from camera frame grabbing till result visualization is provided in algorithm \ref{alg:cap}.

% This method is efficiently implemented using vectorization as shown in the experiments section. 

\begin{algorithm}
\caption{Overall algorithm}\label{alg:cap}
\begin{algorithmic}[1]
\Require Synchronized narrow-angle and wide-angle camera frames, distortion parameters of the two cameras.\\

\textbf{Initialize:} 

\State Estimation of the planar homography $\prescript{w}{}{H_{n}}$ between the two frames after undistortion.
\State Estimation of the region $R_0$ using the homography matrix and distortion parameters.
\State $\mathcal{N}_r$ = $\mathcal{N}_0$

\Repeat
\State Generate $\mathcal{N}$ and $\mathcal{W}$ using the object detector.
\State Generate bounding box set $\mathcal{N}_{0}$ from $\mathcal{N}$. (Fig \ref{fig:bboxtran})
 
\State Obtain $\mathcal{Q}$ and $\mathcal{W}_r$ from region $R_0$ and $\mathcal{W}$. 

\For{$q \in \mathcal{Q}$}

\State Calculate IoU values between $q$ and  all $n_0 \in \mathcal{N}_0$

\If{$\exists$ $n_0 \in \mathcal{N}_0$: IoU $\geq \zeta$}
\State $\mathcal{N}_r$ = $\mathcal{N}_r\backslash\{n_0\}$
\EndIf
\EndFor
\State Output $\mathcal{N}_r \cup \mathcal{W}_r$

\Until{End of camera frame extraction}

\end{algorithmic}
\end{algorithm}

\section{IMPLEMENTATION}\label{implementation}

% \textcolor{red}{Bounding box transformation from wide angle to narrow angel camera did not use undistorted frame}

% \textcolor{red}{Use tensorRT with half precision}

The end-to-end traffic light detection system is implemented as a Robot Operating System (ROS) \cite{ros} package. Synchronized image pair acquisition is carried out using a ROS node programmed using Pylon 6 C++ API \cite{pylon}. For the training and testing of our proposed algorithms, we use a device comprising an Intel Core i9-9900K CPU and a Nvidia RTX-2080 Ti GPU. Additionally, for the purpose of testing the algorithm in a more resource constrained environment, we use Nvidia Jetson AGX Xavier platform.

\subsection{Object detector training and inferencing}

YOLOv5 \cite{yolov5} object detector is provided as 5 different models with varying sizes. We train and evaluate lightweight YOLOv5s \cite{yolov5} model and comparatively large YOLOv5l \cite{yolov5} model to examine how speed-accuracy trade-off affect our algorithm. YOLOv5s \cite{yolov5} is trained for 300 epochs using input size 448$\times$448, batch size 64, learning rate 0.01 and SGD optimizer. YOLOv5l \cite{yolov5} is trained for 300 epochs using input size 640$\times$640, batch size 16, learning rate 0.01 and SGD optimizer. After training we employ half-precision floating point (FP16) TensorRT optimization for batch size two. In algorithm \ref{alg:cap}, 0.5 is used for the value of $\zeta$.

% \hl{Object detectors have an inverse relationship between computational cost and accuracy. When aiming for real-time detection (More than 30 FPS detection speed), accuracy is compromised. In our implementation we aim to improve this limitation by exploiting the high probability of distant traffic lights to be presented in the center of the field of view. Narrow-angle camera captures the distant objects while wide-angle camera captures relatively closer objects. Object detector is given inputs of batch size two, for inferencing. An important point to note is that light weight object detectors can be used for this algorithm while maintaining comparable accuracy. Therefore, though inferencing happens on a batch of size two, significant computational cost can be saved.}

% \textcolor{red}{Include details about the zeta value}

% We train YOLOv5 \cite{yolov5} object detector on our dataset using batch size 64, learning rate of 0.01 and stochastic gradient descent (SGD) optimizer.

\subsection{Homography matrix estimation}\label{homography_estimation}

% In order to calculate the homography between two camera frames, following closed form formula can be used by assuming pinhole camera model. 

% $$
% H_{narrow}^{wide} = k^{wide}\left(\boldsymbol{R} -\boldsymbol{tn}^{T}/d\right)\left( k^{narrow}\right)^{-1} \eqno{(1)}
% $$

% $H_{narrow}^{wide}$ is the homography that maps points in narrow frame to wide frame. $k^{wide}$ and $k^{narrow}$ are the intrinsic camera matrices. $\boldsymbol{R}$, $\boldsymbol{t}$, $\boldsymbol{n}$, $d$ are the rotation matrix, translation vector, plane normal, distance to the plane expressed in narrow-angle camera frame. 

% Due to practical implementation issues, this method does not give accurate matrix. Therefore least-square based estimation using point correspondences is more suitable. To do that a set of corresponding points of interest in two frames is obtained using Scale Invariant Feature Transform (SIFT). \textcolor{red}{ These points belongs to a planar surface which is parallel to the image plane}.
%  % Assuming the points in the world frame is located on a planar surface. 

Planar homography is a mapping between any two images of a planar surface. In order to calculate the homography between two camera frames, following closed form formula can be used by assuming pinhole camera model. 

% $$
% H_{narrow}^{wide} = k^{wide}\left(\boldsymbol{R} -\boldsymbol{tn}^{T}/d\right)\left( k^{narrow}\right)^{-1} \eqno{(1)}
% $$

% \vspace{-0.2cm}

\begin{equation}
\prescript{w}{}{H_{n}} =k_{w}\left(\prescript{w}{}{\mathbf{R}_{n}} -\prescript{w}{}{\boldsymbol{t}_n}\boldsymbol{n}^{T} /d\right) k_{n}^{-1}
\end{equation}
% \vspace{-0.35cm}

$\prescript{w}{}{H_{n}}$ is the homography that maps points in narrow-angle camera frame to wide-angle camera frame. $k_{w}$ and $k_{n}$ are the intrinsic camera matrices. $\prescript{w}{}{\mathbf{R}_{n}}$, $\prescript{w}{}{\boldsymbol{t}_n}$ are the rotation matrix and translation vector between two cameras. $\boldsymbol{n}$, $d$ are the plane normal vector and distance to the chosen plane expressed in narrow-angle camera coordinates.

Due to practical implementation issues, above method does not give an accurate homography matrix. Least square error based estimation using multiple point correspondences is more suitable. Therefore, a set of corresponding points of interest in two frames is obtained using Scale Invariant Feature Transform (SIFT) \cite{lowe2004}. These points belong to a chosen planar calibration surface which is parallel to the image plane.

% Even though we use planar homography to match traffic lights in two frames, it can only be used to match objects on a single plane. Therefore, the traffic lights outside the plane that homography is calculated will not be properly matched. Still, this method gives a low error rate as the narrow angle camera frame fits to a small region in wide angle camera frame and this area can be approximated as a planar area with respect to the rest of the area in the wide-angle camera frame.

Even though we use planar homography to match traffic lights in two frames, it can only be used to match objects on a single plane. Therefore, the traffic lights outside the plane that homography is calculated will not be properly matched. Still, this method gives a lower error rate as the area which maps the narrow-angle camera frame within the wide-angle camera frame $R_0$ can be approximated as a planar area. This is possible, because $R_0$ is quite small and far away with respect to the rest of the area in the wide-angle camera frame.
\section{RESULTS}\label{results}

\subsection{Evaluation metrics}

For the performance evaluation of object detectors on our dataset, we use $F_{1}$-score for each class. Recall is the proportion of correct predictions out of all ground truths. Precision is the proportion of correct predictions out of all predictions. The precision-recall curve is calculated from predictions ranked according to confidence score.

\begin{equation}
Recall = \frac{TP}{TP+FN}
\end{equation}

\begin{equation}
Precision = \frac{TP}{TP+FP}
\end{equation}

\begin{equation}
F1-score = \frac{2 \times Precision \times Recall}{Precision + Recall}
\end{equation}

TP indicates the total number of detected traffic lights (true positives), FN indicates the total number of undetected traffic lights (false negatives) and FP indicates the total number of predictions that cannot be attributed to any ground truth (false positives). A prediction is considered as a true positive based on the IoU value across the ground truth and predicted bounding boxes.

% \begin{equation}
% IoU = \frac{BB_{gt} \cap BB_{detection}}{BB_{gt} \cup BB_{detection}} \geq 0.3
% \end{equation}

\subsection{Dataset evaluation}

For the purpose of evaluating the performance of YOLOv5 \cite{yolov5} on our dataset, inferencing is carried out using batch size one over the test set. We evaluate YOLOv5s \cite{yolov5} and YOLOv5l \cite{yolov5}  with input sizes of 448$\times$448 and 640$\times$640, respectively.

\begin{table}[ht]
\caption{Class-wise $F_{1}$-scores for evaluation on test set}

\begin{center}
\begin{tabular}{|l|c|c|}
% \begin{tabular}{|m{22mm}|p{18mm}|p{18mm}|}
\hline
\cellcolor{gray!15}{\textbf{Class}} & \cellcolor{gray!15}{\textbf{YOLOv5s \cite{yolov5} \newline (FP16)}} & \cellcolor{gray!15}{\textbf{YOLOv5l \cite{yolov5} \newline (FP16)}}\\
\hline
Red               &	72.03	& 83.56 \\ \hline
Green-arrows      &	65.03	& 74.5  \\ \hline
Yellow            &	59.65	& 72.36 \\ \hline
Count-down        &	59.39	& 68.62 \\ \hline
Empty-count-down  &	52.85	& 65.05 \\ \hline
Green             &	49.55	& 62.04 \\ \hline
Red-yellow        &	36.36	& 55.56 \\ \hline
Empty             &	38.25	& 48.59 \\ \hline

\hline
\end{tabular}
\end{center}
\label{table:f1scores}
\end{table}

Table \ref{table:f1scores} compares the class-wise $F_{1}$-scores obtained for the two models using 0.3 as the IoU threshold value. The ground-truth of the test set includes bounding boxes of size reaching a minimum of 6 pixels. YOLOv5l \cite{yolov5} out-performs YOLOv5s \cite{yolov5} in terms of accuracy. An important observation is that some classes seem to perform better compared to others. Count-down, Empty-count-down classes have lower $F_{1}$-scores due to their smaller size compared to traffic lights. Classes such as empty, red-yellow, green-left, green-right and green-up have low number of instances, resulting in lower $F_{1}$-scores. For testing, we combine Green-left, Green-right and Green-up classes to a super class named Green-arrows.

\begin{figure}[!h]
    \centering
    \includegraphics[width=1\columnwidth]{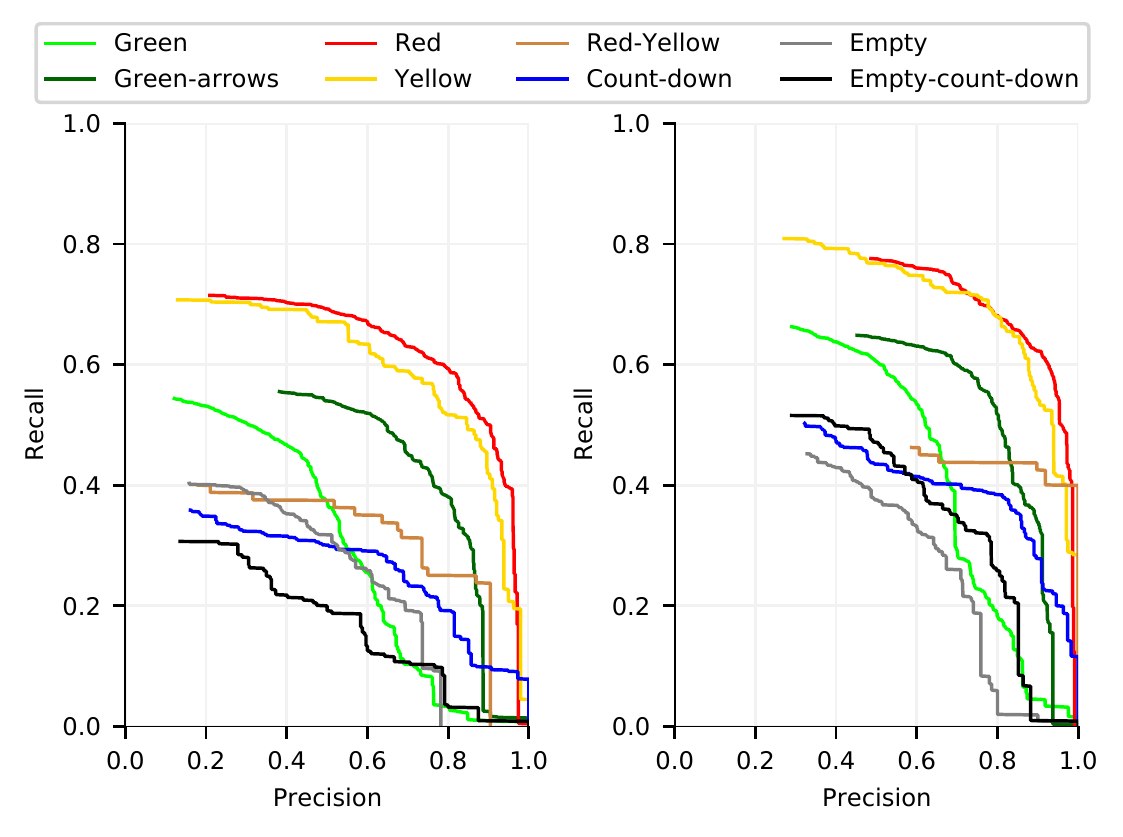}
    \caption{Precision-recall curves of individual classes for YOLOv5s model (left) and YOLOv5l model (right)}
    \label{fig:testpr_new}
\end{figure}

Figure \ref{fig:testpr_new} shows the class-wise precision-recall curves for the two models. In traffic light detection, even though precision can be improved to higher levels, recall has limited improvement capacity \cite{bstld,deeptlr,bach2017}. Clearly, using a larger model such as YOLOv5l \cite{yolov5} can improve predictions by shifting precision-recall curves to the right, compromising on detection speed. 

% as shown in Figure \ref{fig:speed_comparison}

% \begin{figure}[!h]
%     \centering
%      \includegraphics[width=0.5\columnwidth]{images/resultsSt.png}

%     \caption{Precision-Recall curve of individual classes for YOLOv5s model}
%     \label{fig:testpr}
% \end{figure}

% \begin{figure}[!h]
%     \centering
%      \includegraphics[width=0.5\columnwidth]{images/resultsLt.png}

%     \caption{Precision-Recall curve of individual classes for YOLOv5l model}
%     \label{fig:testpr}
% \end{figure}

% 1) Class wise F1 scores

% 2) Precision Recall curves - All classes Comparison

% Models yolov5s , Yolov5l , Yolov4

% \subsection{Synchronized detection}

% Overall ground truth - 2/3 pixel

% compare overall prediction vs overall ground truth

% 1) Zeta value check

% 2) BB distribution for narrow/wide/overall.
\vspace{-0.2cm}
\subsection{Algorithm evaluation}

% speed in Jetson / GPU

% Detection + NMS = milliseconds
% Post processing = Milliseconds
% Bounding box control = milliseconds

\begin{table}[!b]
    \vspace{0.2cm}
    \centering
    \caption{Speed comparison of combined frame approach (pairs per second) with individual frame approach (frames per second).}
    \begin{tabular}{|l|c|c|c|c|} \hline
         \rowcolor{gray!15}\multirow{2}{*}{} & \multicolumn{2}{c|}{\textbf{RTX-2080 Ti}} & \multicolumn{2}{c|}{\textbf{Jetson AGX Xavier}} \\ \hhline{~|----|}
         \rowcolor{gray!15} \raisebox{1.6ex}[1.6ex]{\textbf{Model}} & \textbf{\specialcell{Individual\\frames}}  & \textbf{\specialcell{Combined\\frames}} & \textbf{\specialcell{Individual\\frames}} & \textbf{\specialcell{Combined\\frames}} \\ \hline
        YOLOv5s & 416.7 Hz & 256.4 Hz & 57.5 Hz & 42.8 Hz\\ \hline
        YOLOv5l & 196.1 Hz & 117.6 Hz & 29.5 Hz & 16.5 Hz \\ \hline
    \end{tabular}
    \label{tab:algo_speed_compare}
\end{table}

%  The detection should happen in real time while preserving the high accuracy. Therefore we have evaluated the accuracy and speed of our algorithm using YOLOv5 on our dataset. 

% The overall algorithm achieves a comparable speed compared to single-frame inferencing. 
% Table \ref{tab:algo_speed_compare} compares the inferencing speeds on Nvidia RTX-2080 Ti GPU and Jetson AGX Xavier platform. 

\begin{figure}[t]
    \vspace{0.3cm}
    \centering
    \scriptsize
    \begin{subfigure}[c]{0.49\columnwidth}
         \centering
            \begin{tikzpicture}
            \begin{axis}[
                    % footnotesize,
                    width  = 1.2*\textwidth,
                    height = 4cm,
                    ybar stacked,
                    xtick=data,
                    ymin=0,
                    axis y line*=none,
                    axis x line*=none,
                    y axis line style={->},
                    enlarge x limits=0.5,
                    legend style={at={(0.05,1)},anchor=north west,line width=0.1mm,draw=gray},
                    legend cell align={left},
                    legend entries={batch-1 ,batch-2 ,Post-process},
                    symbolic x coords={YOLOv5s,YOLOv5l},
                    ylabel={Execution time (ms)},
                ]
                \addplot +[bar shift=-.2cm,draw=blue!40, fill=blue!40] coordinates {(YOLOv5s,2.4) (YOLOv5l,5.1) };
                % \addplot +[bar shift=-.2cm] coordinates {(1,0) (2,0)};
                
                \resetstackedplots
                
                \addplot  +[bar shift=.2cm,draw=red!45,fill=red!45]coordinates {(YOLOv5s,2.9) (YOLOv5l,7.7) };
                \addplot  +[bar shift=.2cm ,fill=brown!45, draw=brown!45] coordinates {(YOLOv5s,1) (YOLOv5l,0.8) };
            
                \end{axis}
            \end{tikzpicture}
         \caption{}
        %  \label{fig:y equals x}
        % \vspace*{1mm}
     \end{subfigure}
     \hfill
    \begin{subfigure}[c]{0.49\columnwidth}
         \centering
         \begin{tikzpicture}
            \begin{axis}[
                    % footnotesize,
                    width  = 1.2*\textwidth,
                    height = 4cm,
                    ybar stacked,
                    xtick=data,
                    ymin=0,
                    axis y line*=none,
                    axis x line*=none,
                    y axis line style={->},
                    enlarge x limits=0.5,
                    symbolic x coords={YOLOv5s,YOLOv5l},
                ]
                \addplot +[bar shift=-.2cm, draw=blue!40, fill=blue!40] coordinates {(YOLOv5s,17.4) (YOLOv5l,33.9) };
                % \addplot +[bar shift=-.2cm] coordinates {(1,0) (2,0)};
                
                \resetstackedplots
                
                \addplot  +[bar shift=.2cm, draw=red!45,fill=red!45]coordinates {(YOLOv5s,18.3) (YOLOv5l,55.5) };
                \addplot  +[bar shift=.2cm, draw=brown!45,fill=brown!45] coordinates {(YOLOv5s,5) (YOLOv5l,5) };
            
                \end{axis}
            \end{tikzpicture}
         \caption{}
        %  \label{fig:y equals x}
        % \vspace*{0.1mm}
     \end{subfigure}

    \caption{Speed comparison (a) on RTX-2080 Ti GPU (b) on Jetson AGX Xavier platform. Batch-1 and batch-2 denotes inferencing with batch size 1 and 2, respectively.}
    \label{fig:speed_comparison}
    \vspace{-0.3cm}
\end{figure}
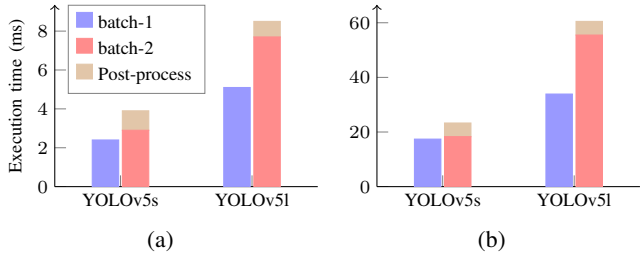

Our proposed algorithm \ref{alg:cap} includes batch size 2 inferencing followed by a post processing algorithm. The speed of the combined frame approach is measured by considering the time spent on processing a pair of frames. This is compared with frame rate in individual frame approach. Table \ref{tab:algo_speed_compare} shows the speed comparison for Nvidia RTX-2080 Ti GPU and Jetson AGX Xavier platform. Combined frame approach achieves more than half the speed achieved by the single frame approach. Even though Jetson AGX Xavier has constrained resources, it performs in real-time when YOLOv5s \cite{yolov5} is used. However, it only achieves a speed of 16 Hz for the case of YOLOv5l \cite{yolov5}. The post processing algorithm takes up additional 1ms computation time in RTX-2080 Ti GPU whereas it takes up 5ms in Jetson AGX Xavier with minimum impact on overall speed as shown in the Figure \ref{fig:speed_comparison}.

% Jetson AGX Xavier takes upto 5ms with minimum impact on overall speed as shown in the Figure \ref{fig:speed_comparison}.

% In our approach the speed has been \textcolor{red}{reduced by 30\%} in the GPU and \textcolor{red}{xx\%} in Jetson AGX Xavier board. 
%%%%%%%%%%%%%%%%5

% Our main objective is to detect the traffic lights even when they are far away from the vehicle while detecting the nearby traffic lights as well.

% The positive effect of combining bounding boxes from two frames is shown in Figure \ref{fig:ablation}. 

\begin{figure}[!h]
    \centering
     \includegraphics[width = 1\columnwidth]{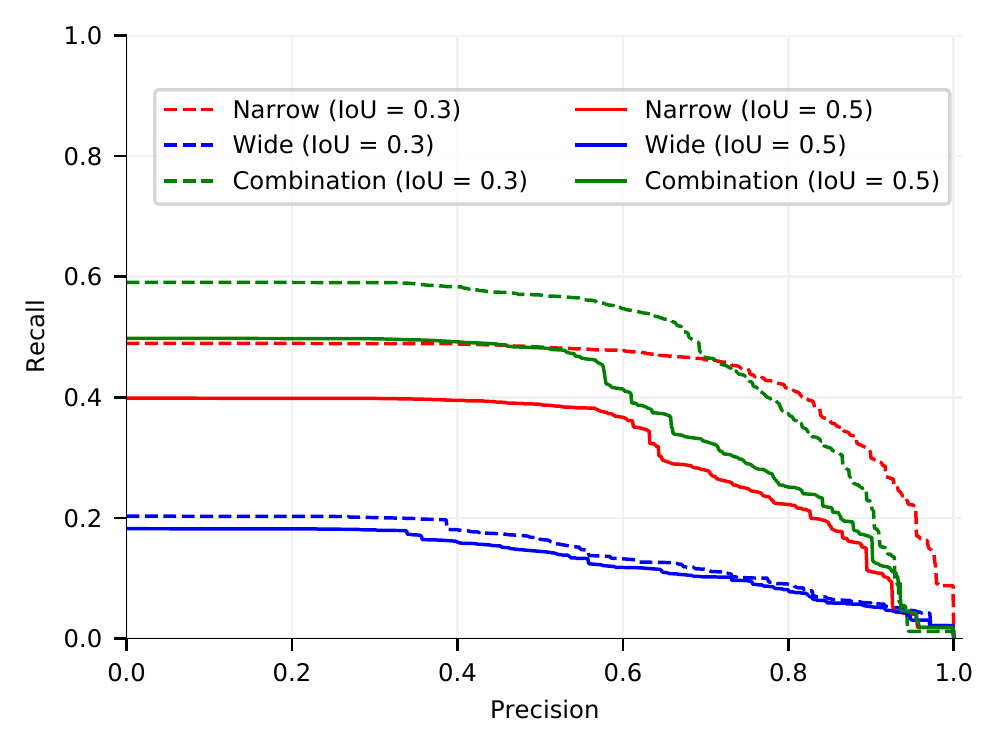}
    \caption{Overall precision-recall curves with respect to common ground truth. Detection using combination of both frames out-performs the detection using individual frames.}
    \label{fig:ablation}
\end{figure}

Reduction in small amount of speed leads to a significant improvement in accuracy by using our technique as shown in Figure \ref{fig:ablation}. The precision-recall curves are generated using a common ground-truth prepared using synchronized images in the test set. This ground-truth contains bounding boxes of size reaching a minimum of 1 pixel. The resulting combined bounding boxes from the algorithm \ref{alg:cap} achieves higher recall values compared to individual contributions from narrow-angle and wide-angle camera frames. The recall values reach higher values when the IoU threshold is 0.3 compared to the IoU threshold of 0.5. The contribution of narrow-angle camera is higher compared to the wide-angle camera. This is due to the prevalence of higher number of instances in the narrow-angle camera frames.

%%%%%%%%%%%%%%%%%%%%%%%%

\vspace{-0.3cm}

\vspace{-0.3cm}
\section{CONCLUSION}

This paper introduced DualCam, a novel traffic light benchmark dataset, addressing the need for synchronized images captured using a pair of cameras.
The proposed post-processing algorithm can efficiently combine the detection output from the two cameras with minimum impact on the overall detection algorithm. Results show that our approach results in a considerable increase in recall, with a large contribution from the narrow-angle camera, compared to the conventionally used wide-angle cameras. Incorporating multiple cameras and assigning traffic lights to respective lanes are possible future extensions of the work presented in this paper.

% Evaluation results show that narrow-angle and wide-angle camera output combination with proper post-processing results in a fair compromise between accuracy and speed.

% An algorithm is introduced for overcoming the challenges of using narrow and wide angle cameras together for real-time detection. 

\addtolength{\textheight}{-1cm}   % This command serves to balance the column lengths
                                  % on the last page of the document manually. It shortens
                                  % the textheight of the last page by a suitable amount.
                                  % This command does not take effect until the next page
                                  % so it should come on the page before the last. Make
                                  % sure that you do not shorten the textheight too much.

%%%%%%%%%%%%%%%%%%%%%%%%%%%%%%%%%%%%%%%%%%%%%%%%%%%%%%%%%%%%%%%%%%%%%%%%%%%%%%%%

%%%%%%%%%%%%%%%%%%%%%%%%%%%%%%%%%%%%%%%%%%%%%%%%%%%%%%%%%%%%%%%%%%%%%%%%%%%%%%%%

%%%%%%%%%%%%%%%%%%%%%%%%%%%%%%%%%%%%%%%%%%%%%%%%%%%%%%%%%%%%%%%%%%%%%%%%%%%%%%%%
% \section*{APPENDIX}

% Appendixes should appear before the acknowledgment.

% \section*{ACKNOWLEDGMENT}

% Use if needed.

%%%%%%%%%%%%%%%%%%%%%%%%%%%%%%%%%%%%%%%%%%%%%%%%%%%%%%%%%%%%%%%%%%%%%%%%%%%%%%%%

% \subfile{sections/references}
\bibliographystyle{IEEEtran}
\bibliography{references}

\end{document}